\documentclass[sigconf]{acmart}
\AtBeginDocument{%
  }

\usepackage[utf8]{inputenc}

\usepackage{booktabs}
\usepackage{multirow}
\usepackage{graphicx}

\copyrightyear{2027}
\acmYear{2027}
\setcopyright{cc}
\setcctype{by}
\acmConference[ASPDAC '27]{32nd Asia and South Pacific Design Automation Conference}{January 25--28, 2027}{Tokyo, Japan}
\acmBooktitle{32nd Asia and South Pacific Design Automation Conference (ASPDAC '27), January 25--28, 2027, Tokyo, Japan}
\acmDOI{10.1145/3840404.3848401}
\acmISBN{979-8-4007-2923-2/2027/01}

\usepackage{graphicx}

\usepackage{booktabs,multirow,amsmath,graphicx}

\usepackage[acronym,ucmark,translate]{glossaries}
\newacronym{ai}{AI}{Artificial Intelligence}
\newacronym{bnpa}{BNPA}{Bottleneck Normalized Projected Adapters}
\newacronym{cnn}{CNN}{Convolutional Neural Network}
\newacronym{iot}{IoT}{Internet of Things}
\newacronym{lora}{LoRA}{Low-Rank Adaptation}
\newacronym{llm}{LLM}{Large Language Model}
\newacronym{mac}{MAC}{Multiply-Accumulate}
\newacronym{nn}{NN}{Neural Network}
\newacronym{peft}{PEFT}{parameter-efficient fine-tuning}
\newacronym{sg}{SG}{streaming-gradient}
\newacronym{sram}{SRAM}{Static Random Access Memory}
\newacronym{har}{HAR}{Human Activity Recognition}
\newacronym{memflora}{MemFLoRA}{Memory-Floor LoRA}
\newacronym{loso}{LOSO}{Leave-One-Subject-Out}
\newacronym{lolo}{LOLO}{Leave-One-Location-Out}
\newglossary[slg]{symbols}{syi}{syg}{List of symbols} 
\begin{document}

\title{MemFLoRA: Memory-Floor LoRA for CNN Adaptation at the Edge}

\author{Mehmet Emre Akbulut}
\orcid{0009-0007-7372-2580}
\affiliation{%
  \institution{Technical University of Munich}
  \city{Munich}
  \country{Germany}
}
\email{mehmet.akbulut@tum.de}

\author{Johannes Geier}
\orcid{0000-0002-9439-3890}
\affiliation{%
  \institution{Technical University of Munich}
  \city{Munich}
  \country{Germany}
}
\email{johannes.geier@tum.de}

\author{Ulf Schlichtmann}
\orcid{0000-0003-4431-7619}
\affiliation{%
  \institution{Technical University of Munich}
  \city{Munich}
  \country{Germany}
}
\email{ulf.schlichtmann@tum.de
}



\begin{abstract}

On-device learning is necessary when the model encounters user-, sensor-, or environment-specific shifts after deployment.
Although \gls{peft} methods, particularly \gls{lora} variants, enable efficient adaptation at the edge, the limiting resource for \gls{cnn} adaptation is often not the number of trainable parameters but the activation state that must be retained until the backward pass.
This paper introduces \gls{memflora}, a low-rank \gls{cnn} adapter built around a memory-first design principle rather than a direct application of transformer-oriented LoRA.
Instead of merely reducing trainable weights, we define an activation-memory-floor criterion: trainable backward computations must not depend on full-width layer inputs.
The resulting adapter freezes the down-projection, trains a scale-matched up-projection, and combines eval-mode backbone normalization with activation-minimal backward rules, reducing saved state to the low-rank branch.

Evaluated on three \gls{har} datasets and two CNN backbones under subject, body-location, and sensor-placement shifts, MemFLoRA reduces saved-activation memory by $98.5$-$98.7\%$ and peak training-state memory by $94.9$-$97.3\%$ relative to full
fine-tuning, while matching or exceeding CNN PEFT baselines.


\end{abstract}




\maketitle

\section{Introduction}
\Gls{ai} at the edge has become increasingly important as intelligent sensing devices become deeply embedded in everyday life. Smartphones, wearables, and \gls{iot} platforms interact with their environments and generate large amounts of domain-specific data. This trend pushes \gls{ai} computation closer to the data source, enabling lower latency, reduced communication costs, and improved privacy compared with sending all data to cloud servers \cite{10.1145/3613424.3614307}.

\begin{figure}
    \centering
    \includegraphics[width=\linewidth]{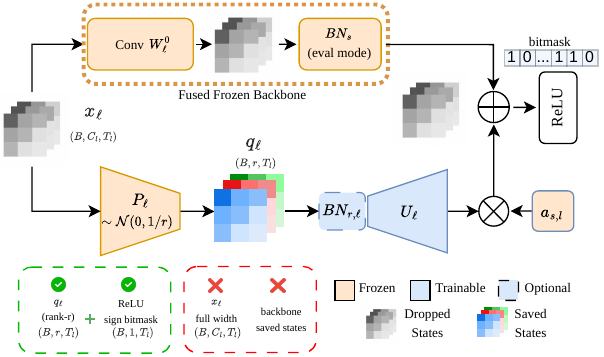}
    \caption{Brief overview of MemFLoRA computation and saved-state graph, reducing persistent saved state from $O(B\,C_\ell\,T_\ell)$ to $O(B\,r\,T_\ell)$ plus a bit-packed ReLU mask.}
    \label{fig:main_diagram}
    \Description[MemFLoRA computation and saved-state flow]{
MemFLoRA computation and saved-state flow. An input feature tensor
splits into a frozen backbone branch and a low-rank adapter branch.
The backbone branch applies a frozen convolution followed by
evaluation-mode batch normalization. The adapter branch applies a
frozen pointwise down-projection initialized from a normal
distribution, optionally applies bottleneck batch normalization,
then applies a trainable up-projection and a fixed per-channel scale
inherited from the source batch normalization. The two branches are
added and passed through a rectified linear unit. The diagram shows
that only the bottleneck activation, whose channel count equals the
adapter rank, and a bit-packed rectified-linear-unit sign mask are
retained. The full-width input, backbone activations, and
batch-normalization states are dropped rather than persistently
saved for backward computation.
}
\end{figure}

However, a fixed pretrained model may fail to provide robust predictions after deployment. For example, in sensor-based \gls{har}, the target stream can differ from the source data due to subject-dependent motion patterns, sensor-placement deviations, device heterogeneity, and evolving user behavior~\cite{10.1109/MM.2023.3314711, NADALINI2023212}.
These changes can lead to distribution shifts or concept drifts and degrade the out-of-distribution performance of fixed pretrained models.
Since the corrective data are generated locally and may be privacy-sensitive, adapting the model on-device is preferable to repeated cloud-side retraining. However, full fine-tuning updates all model parameters and incurs substantial memory, compute, and energy overhead, which can exceed the resources of embedded systems and motivate more selective adaptation strategies.

Since its introduction~\cite{Hu2021LoRALA}, \gls{lora}  has become a widely used adaptation method for \glspl{llm}, but its role in \glspl{cnn} and extreme-edge learning scenarios remains much less explored. The standard reading of \gls{peft} is that reducing the number of trainable parameters reduces the training-memory footprint. This assumption is often reasonable in \glspl{llm}, where optimizer state and trainable weight gradients dominate. 
However, for edge \glspl{cnn}, memory is often dominated by the activation tensors that automatic differentiation retains for backward propagation, not by the optimizer state or the trainable parameter set~\cite{NEURIPS2020_81f7acab}.

This mismatch indicates that parameter-efficient methods may still be memory-expensive, motivating MemFLoRA\footnote{Code is available at
\url{https://github.com/mehmetemreakbulut/MemFLoRA}.}: a \gls{lora}-style \gls{cnn} adapter with an activation-memory floor. The design rule is that no trainable backward computation should require the full-width activation tensor.
The adapter satisfies this rule through a frozen down-projection, a trainable up-projection scale-matched to the frozen backbone, and eval-mode backbone normalization.
This restricts the saved state to the low-rank branch while preserving adaptation capacity.
In summary, our main contributions are:

\begin{itemize}
    \item By decomposing the saved-tensor cost into three pillars, we define an activation-memory-floor criterion requiring that no persistent full-precision activation tensor used by trainable backward computations scales with layer width, and show that MemFLoRA satisfies it with rank-$r$ saved state. Enforcing this floor reduces saved-activation memory by $98.5$-$98.7\%$ and total peak training-state memory by $94.9$-$97.3\%$ relative to full fine-tuning.
    \item We introduce a memory-floor \gls{cnn} adapter that combines a frozen down-projection, a scale-matched trainable {up-pro}-jection, eval-mode backbone normalization, and activation-minimal backward rules, together with a \gls{sg} variant that recovers the exact projection gradient without persistently saving the full-width activation.
    \item Across three \gls{har} datasets spanning subject, location, and sensor-placement shifts and two \gls{cnn} backbones, we match or exceed all evaluated rank-dependent PEFT methods in mean Macro-F1 across ranks and backbones.
\end{itemize}

\section{Background and Related Work}

Prior work on efficient adaptation has followed two complementary directions. \gls{lora}-style \gls{peft} methods restrict updates to low-rank branches, reducing trainable parameters and optimizer-state costs while remaining easy to integrate into pretrained models and, in many cases, mergeable into the backbone for inference. On-device training methods instead prioritize the dominant training-memory cost by reducing saved activations through selective updates and freezing. Our work bridges these directions by co-designing the adapter architecture and training graph around this activation-memory constraint, rather than treating activation reduction and adaptation as independent problems.

\subsection{Low-Rank Adaptation}

\Gls{lora} introduced low-rank adaptation by injecting trainable low-rank updates into frozen pretrained weights, reducing trainable parameters while preserving accuracy \cite{Hu2021LoRALA}. Most low-rank adaptation research focuses on transformer-based models and the adapter architecture. Work on \glspl{cnn} is sparser and largely adopts the same parameter-centric perspective as its transformer-based counterparts ~\cite{Ding2024LoRACPF,kwak2025loraedge,10.1145/3658617.3697589} rather than reconsidering the activation state retained during training. ConvLoRA \cite{10635661} combines \gls{lora} with AdaBN~\cite{LI2018109} for medical-image domain adaptation, whereas Convolution Meets LoRA \cite{Zhong2024ConvolutionML} integrates lightweight convolutions into \gls{lora} experts without fine-tuning the convolution itself. LoRA-C~\cite{Ding2024LoRACPF} performs low-rank decomposition at the
convolutional-layer level rather than the kernel level, targeting
parameter-efficient adaptation for IoT devices. More recently, LoRA-Edge~\cite{kwak2025loraedge} targets on-device \gls{cnn} fine-tuning for human activity recognition under domain shift, using a tensor-train structure to reduce trainable parameters and improve convergence.
Across this body of work, the shared objective is parameter efficiency, leaving the \gls{cnn} setting's dominant saved-activation cost largely unaddressed.

LoRA-FA ~\cite{zhang2023lorafa} freezes the down-projection and trains only the up-projection. Theoretical work on asymmetric low-rank adapters further explains that freezing the input-side projection constrains updates to a projected subspace, while leaving the output-side factor trainable can remain expressive in high-dimensional regimes \cite{3692070.3694651}. MemFLoRA inherits this asymmetric-trainability principle but extends it to \gls{cnn} adaptation, where BatchNorm, activation backward, and convolutional execution introduce saved-state constraints absent from
the transformer setting. Asymmetric trainability alone removes the full-width input saved for the trainable down-projection. In \glspl{cnn}, however, full-width state can also be retained by train-mode BatchNorm, activation backward, and the frozen convolutional branch.

\subsection{On-Device Training and Activation-Memory}

Several prior systems reduce training memory for on-device learning through quantization, sparse updates, rematerialization, or activation compression. TinyTL~\cite{NEURIPS2020_81f7acab} showed that activations dominate on-device training memory. Building on this perspective, MCUNetV3 \cite{Lin2022OnDeviceTU} combines quantization-aware scaling, sparse updates, and a tiny training engine to enable learning under extremely small memory budgets. POET \cite{Patil2022POETTN} formulates rematerialization and paging schedules for training, while gradient checkpointing \cite{Chen2016TrainingDN} reduces activation storage by recomputing intermediate activations during the backward pass. Activation-compression methods such as ActNN~\cite{pmlr-v139-chen21z}, GACT~\cite{Liu2022GACTAC}, and Few-bit Backward~\cite{Novikov2022FewBitBQ} reduce the cost of storing activations by compressing or quantizing them.

Checkpointing, rematerialization, and activation compression are
orthogonal execution-level techniques: they reduce or reconstruct
the state required by a given adapter but do not change its intrinsic
backward dependencies. Our floor instead concerns adapter design,
before such techniques are applied. MemFLoRA attains this floor
without recomputation or compression, and these techniques can be
applied on top of it for further savings.

MemFLoRA combines asymmetric low-rank adaptation with the
activation-centric perspective of on-device training. Unlike
the execution-level techniques discussed above,
it modifies the adapter's intrinsic backward dependencies so that
the persistent state required by trainable computations is bounded
by the rank-$r$ branch. To the best of our knowledge, MemFLoRA is
the first LoRA-style CNN adapter explicitly designed around an
activation-memory-floor criterion. The novelty lies in enforcing a graph-level activation-memory constraint across the frozen projection, adapter branch, backbone normalization, activation backward, adapter placement, convolutional geometry, and optionally, projection-gradient recovery.

\section{Method}
\label{sec:method}
 
We introduce a family of LoRA-style adapters for activation-memory-efficient CNN adaptation as illustrated in Figure~\ref{fig:main_diagram}. In
transformers, \gls{lora} is naturally applied to linear projections,
where the main design variable is the rank of the injected update. In
\glspl{cnn}, however, the adapter interacts with convolutional kernel
geometry, BatchNorm mode and statistics, activation functions, and
fused Conv--BN--ReLU execution. Consequently, directly factorizing
convolutional weights may reduce trainable parameters while still
retaining full-width feature maps for the backward pass. We instead require that no trainable backward formula should require the full-width activation tensor.
In the remainder of this section, we define the memory floor, derive the adapter form and placement from it, and show how each component follows from this constraint rather than being an independent design choice.

\subsection{The Activation-Memory Floor Criterion}
\label{sec:floor}

\begin{figure}[t]
    \centering
    \includegraphics[width=1\linewidth]{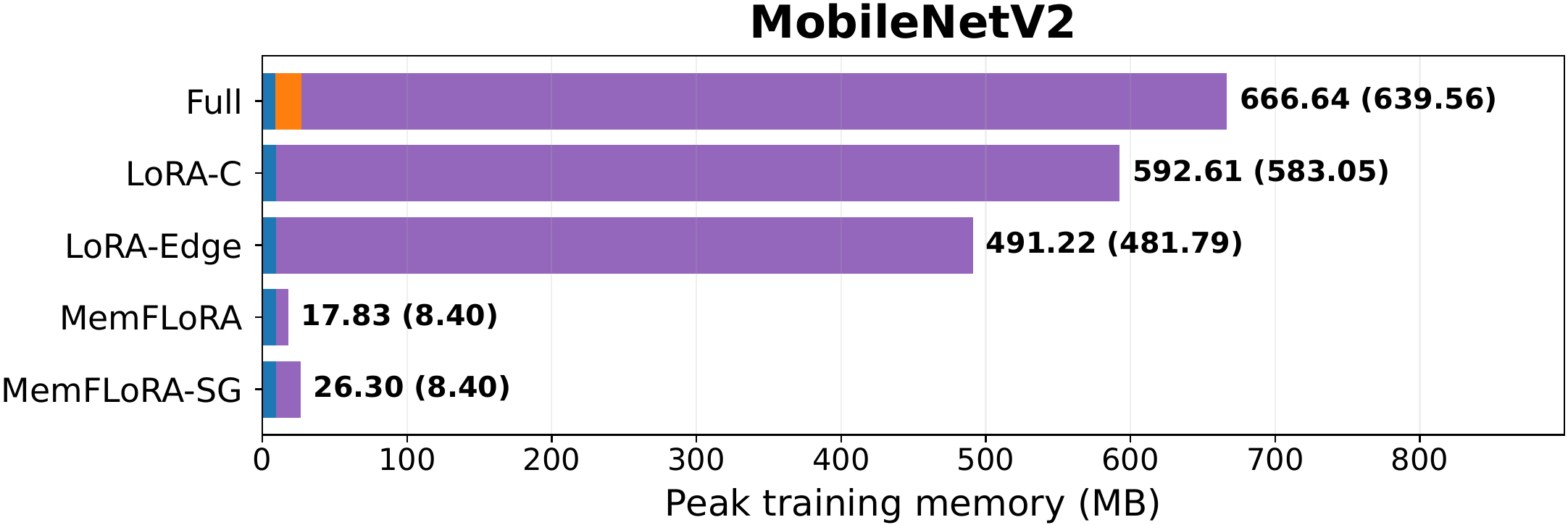}
    \caption{
    Peak training-state memory decomposition of MobileNetV2 at rank-2 on Opportunity with a batch size of 64. Labels report total peak memory, with saved-activation memory in parentheses.
    }
   
    \label{fig:memory_story}
     \Description[MobileNetV2 peak training-memory decomposition]{
Stacked horizontal bars compare peak training memory for MobileNetV2
on Opportunity at rank two and batch size sixty-four. The five bars
represent Full Fine-Tuning, LoRA-C, LoRA-Edge, MemFLoRA, and
MemFLoRA-SG. Each bar is divided into model state, optimizer state,
and live training state. Live training state occupies most of the
memory for Full Fine-Tuning, LoRA-C, and LoRA-Edge. Total peak memory
and saved-activation memory, both in megabytes, are 666.64 and 639.56
for Full Fine-Tuning, 592.61 and 583.05 for LoRA-C, 491.22 and 481.79
for LoRA-Edge, 17.83 and 8.40 for MemFLoRA, and 26.30 and 8.40 for
MemFLoRA-SG. The MemFLoRA bars are substantially shorter than the
three conventional training methods, while MemFLoRA-SG has a
slightly larger total than MemFLoRA because of its additional replay
state.
}
\end{figure}

Total training memory decomposes as
\begin{equation}
M_{\mathrm{train}}
=
M_{\mathrm{weights}}
+ M_{\mathrm{grad}}
+ M_{\mathrm{optim}}
+ M_{\mathrm{saved}}
+ M_{\mathrm{buf}},
\label{eq:mtrain}
\end{equation}
where the terms denote model weights, gradients, optimizer states, tensors saved during the forward pass for backward reuse, and temporary buffers, respectively. Among these, $M_{\mathrm{saved}}$
can dominate for edge \glspl{cnn} (Figure~\ref{fig:memory_story}). 
Parameter-efficient fine-tuning reduces $M_{\mathrm{grad}}$ and $M_{\mathrm{optim}}$ but leaves $M_{\mathrm{saved}}$ essentially unchanged, which is why parameter efficiency does not imply memory efficiency.

For input $x_\ell \in \mathbb{R}^{B \times C_\ell \times T_\ell}$, $M_{\mathrm{saved}}$ decomposes into three \emph{pillars}, each retained by a distinct part of the backward graph: 
\emph{(i) activation/ReLU state}: although ReLU backward requires only the activation sign, standard automatic differentiation typically retains a full-width tensor,
$O(B\,C_\ell\,T_\ell)$;
\emph{(ii) BatchNorm state}: train-mode BatchNorm's backward requires the normalized input, so full-width BN input is saved, $O(B\,C_\ell\,T_\ell)$;
\emph{(iii) frozen-branch convolution state}: the input of every convolution is saved, including the input of the adapter and the input of the frozen backbone.
A standard low-rank convolutional adapter trains both factors and therefore its saved adapter input remains full-width, $O(B\,C_\ell\,T_\ell)$, matching the asymptotic activation width of full fine-tuning.


We define the activation-memory floor for single-pass LoRA-style
CNN adapters as persistent saved state of
$O(B r T_\ell)$ for trainable backward computations,
independent of $C_\ell$. The ReLU gate is accounted for separately
as a bit-packed mask, which is the minimal direct state for exact ReLU backward
without recomputation.
MemFLoRA satisfies this by replacing the dense saved state
of the three pillars with the rank-$r$ bottleneck and a bit-packed
ReLU gate.
\\
\textbf{Eval-mode BatchNorm} removes Pillar~(ii): During adaptation, the BatchNorm statistics $(\mu_\ell,\sigma_\ell^2)$ are frozen to target-calibrated values, and the affine parameters $(\gamma_{s,\ell}, \beta_{s,\ell})$ are the fixed backbone scale and shift. BatchNorm then reduces to a per-channel affine map

\begingroup
\setlength{\abovedisplayskip}{2pt}
\setlength{\belowdisplayskip}{2pt}
\begin{equation}
\mathrm{BN}^{\mathrm{eval}}_s(z)
=a_{s,\ell}\odot z+b_{s,\ell};\;
a_{s,\ell}=\tfrac{\gamma_{s,\ell}}
{\sqrt{\sigma_{s,\ell}^{2}+\varepsilon}};\;
b_{s,\ell}=\beta_{s,\ell}-a_{s,\ell}\odot\mu_{s,\ell}
\label{eq:bnaffine}
\end{equation}
\endgroup

Because ~\eqref{eq:bnaffine} is a fixed affine transform, its backward pass
saves no activation. Eval-mode BatchNorm is therefore a precondition for the floor rather than an optional choice. 
\\
\textbf{Bit-packed sign mask} reduces Pillar~(i): the backward pass of a fused Conv--BN--ReLU operator stores the ReLU sign at one bit per element rather than a full-precision tensor.
\\
\textbf{Frozen down-projection} removes Pillar~(iii): We adapt each convolution with a frozen down-projection $P_\ell$ and a trainable up-projection $U_\ell$.
If $P_\ell$ were trainable, its gradient has the outer-product form
$
\nabla P_\ell \;\propto\; g_\ell\, x_\ell^\top
\label{eq:gradP}
$
which would require retaining the full-width input $x_\ell$, of size $O(B\,C_\ell\,T_\ell)$, violating the floor. After freezing $P_\ell$, $U_\ell$ requires only the bottleneck activation
$
q_\ell = P_\ell\, x_\ell \in \mathbb{R}^{B \times r \times T_\ell}
\label{eq:bottleneck}
$
reducing the saved adapter state to $O(B\,r\,T_\ell)$. At $r=1$,
this becomes $O(B\,T_\ell)$ and is independent of $C_\ell$.
Frozen-projection asymmetry is thus a requirement of the floor, not a stylistic choice. 

\subsection{Adapter Form and Placement}
\label{sec:adapter}

We adapt each convolution with a branch consisting of a frozen down-projection $P_\ell$ and a trainable up-projection $U_\ell$:
\begin{equation}
\Delta_\ell = U_\ell\big(P_\ell\, x_\ell\big) = U_\ell(q_\ell),
\end{equation}

As shown in Section~\ref{sec:floor}, freezing $P_\ell$ places this branch at the memory-floor: the gradient of $U_\ell$ depends only on the rank-$r$ bottleneck $q_\ell$, never on the full-width input $x_\ell$.

To preserve the activation-memory floor, we implement each Conv--BN--ReLU site as a single activation-minimal operator rather than exposing the convolution output, BN output, and ReLU preactivation as separate saved tensors. 
Under this memory constraint, the adapter branch is added to the fused Conv--BN output:
\begin{equation}
y_\ell
= \varphi\!\left(
\mathrm{BN}_s^{\mathrm{eval}}\!\big(W^0_\ell\, x_\ell\big)
+ 
\Delta_\ell
\right)
\label{eq:forward}
\end{equation}

Here, $W_\ell^0$ is the frozen source convolution,
$\varphi$ is the activation function, and $\Delta_\ell$ is the
low-rank adapter update.

Because eval-mode BatchNorm is affine
(Equation~\eqref{eq:bnaffine}), injecting the adapter before
normalization is algebraically equivalent to
\begin{equation}
\begin{aligned}
&\varphi\!\left(
\mathrm{BN}_s^{\mathrm{eval}}
(W_\ell^0x_\ell+\Delta_\ell)
\right) =
\varphi\!\left(
\mathrm{BN}_s^{\mathrm{eval}}(W_\ell^0x_\ell)
+a_{s,\ell}\odot\Delta_\ell
\right).
\end{aligned}
\label{eq:equiv}
\end{equation}

\subsection{Scale-Matching the Adapter Branch}
\label{sec:scale}

The forward rule in ~\eqref{eq:forward} adds the adapter output $\Delta_\ell$ to the normalized backbone output $\mathrm{BN}_s^{\mathrm{eval}}(W^0_\ell x_\ell)$.  The backbone term is scaled per channel by $a_{s,\ell}$ per~\eqref{eq:bnaffine}; the adapter term, at raw scale, is not.  This subsection shows that matching the per-channel scale of the two branches is critical under the evaluated initialization and optimization protocol, examines the two ways to supply it, and derives our choice. Writing the pre-activation as
\begin{equation}
z_\ell
= a_{s,\ell}\odot(W^0_\ell x_\ell) + b_{s,\ell}
+ 
U_\ell(q_\ell),
\label{eq:preact}
\end{equation}
the bottleneck $q_\ell = P_\ell x_\ell$ is a frozen random projection of the layer input; its per-channel magnitudes are uncontrolled and are not on the scale of the normalized backbone term. 
When the two branches are mismatched in scale, the adapter updates are poorly
conditioned across channels, the adapter output rapidly grows out of scale with the
normalized backbone term, and the ReLU gate destabilizes. 
Empirically, a raw-scale memory-floor adapter fails to adapt at all (Section~\ref{sec:results}, Table~\ref{tab:ablation_design_choices}). The adapter branch, therefore, requires per-channel \emph{scale-matching} to the backbone before summation.

\paragraph{Option A: trainable bottleneck normalization.}
One way to supply the missing scale is to normalize the bottleneck before $U_\ell$ with a trainable BatchNorm $\mathrm{BN}_{r,\ell}$ over the $r$
bottleneck channels,
\begin{equation}
\Delta_\ell^{\mathrm{BN}_r}
= U_\ell\!\ \bigl(
\gamma_{r,\ell} \odot
\frac{q_\ell-\mu_{r,\ell}}
{\sqrt{\sigma_{r,\ell}^2+\varepsilon}}
+\beta_{r,\ell}
\bigr),
\label{eq:bnr}
\end{equation}
with batch statistics $(\mu_{r,\ell}, \sigma_{r,\ell}^2)$ and trainable affine $(\gamma_{r,\ell}, \beta_{r,\ell})$.
This normalization places every bottleneck channel at unit scale and prevents the collapse of ~\eqref{eq:preact}.
Batch-dependent statistics add variation under small adaptation batches and discard the per-channel bottleneck magnitudes used by the trainable spatial kernel.

\paragraph{Option B: inherited scale.}
The equivalence in ~\eqref{eq:equiv} identifies a \emph{natural}
per-channel scale directly: it is the eval-mode BatchNorm output scale
$a_{s,\ell}$, the same constant that the fused block already applies to the backbone
branch. Scaling the adapter output by $a_{s,\ell}$ yields
\begin{equation}
\;
y_\ell
= \varphi\!\left(
\mathrm{BN}_s^{\mathrm{eval}}\!\big(W^0_\ell x_\ell\big)
+ a_{s,\ell} \odot U_\ell(q_\ell)
\right)
\;
\label{eq:final}
\end{equation}
Unlike $\mathrm{BN}_{r,\ell}$, the factor $a_{s,\ell}$ is a fixed, precomputed, per-channel constant, so it introduces no batch-statistic noise and no additional saved state, and it leaves the bottleneck magnitudes intact for $U_\ell$.

\paragraph{Choice.}
Under the matched initialization and geometry presented in Section \ref{sec:geometry}, the fixed scale $a_{s,\ell}$ outperforms the trainable normalization $\mathrm{BN}_{r,\ell}$ (Table~\ref{tab:ablation_design_choices}): both prevent the collapse, but $a_{s,\ell}$ avoids the statistical noise caused by $\mathrm{BN}_{r,\ell}$ and preserves the bottleneck signal.
Although $\mathrm{BN}_{r,\ell}$ is initialization-agnostic and mitigates the effect of projection, we adopt ~\eqref{eq:final} and use no bottleneck normalization.

\subsection{Convolutional Geometry}
\label{sec:geometry}

The adapter is defined by a frozen down-projection $P_\ell$ and a trainable up-projection $U_\ell$, but it does not yet specify \emph{which factor carries the spatial kernel}.
For a linear layer, this question does not arise; for a convolution, it is a design choice determining whether the adapter can re-learn temporal or spatial filters for the target domain.
Two parameterizations are on the memory floor, since both keep the bottleneck $q_\ell$ at rank $r$.

\paragraph{Frozen-kernel parameterization.}
The down-projection $P_\ell$ carries the $k$-tap spatial kernel and is frozen;
the trainable $U_\ell$ is a pointwise ($1\times1$) channel mixer:
\begin{equation}
P_\ell \in \mathbb{R}^{r \times C_\ell \times k}, \quad
U_\ell \in \mathbb{R}^{C_{\mathrm{out}} \times r \times 1}, \quad
\#(U_\ell) = r\,C_{\mathrm{out}}.
\label{eq:geomfrozen}
\end{equation}
Here, the spatial filters are fixed, so the adapter can only re-mix a fixed set of filter responses; it cannot form new temporal filters.

\paragraph{Trainable-kernel parameterization.}
The down-projection is a frozen pointwise projection and the trainable up-projection carries the full $k$-tap spatial kernel:
\begin{equation}
P_\ell \in \mathbb{R}^{r \times C_\ell \times 1}, \quad
U_\ell \in \mathbb{R}^{C_{\mathrm{out}} \times r \times k}, \quad
\#(U_\ell) = r\,C_{\mathrm{out}}\,k.
\label{eq:geomours}
\end{equation}
The trainable spatial kernel operates directly on the bottleneck magnitudes $q_\ell$, which are the same magnitudes that Section~\ref{sec:scale} showed must be
preserved rather than normalized away. 

\paragraph{Parameter cost and the floor.}
The trainable-kernel parameterization costs a factor of $k$ in trainable parameters, $r\,C_{\mathrm{out}}\,k$ versus $r\,C_{\mathrm{out}}$.
This cost falls on the optimizer state $M_{\mathrm{optim}}$, not on the activation floor:
the saved adapter state is $O(B\,r\,T_\ell)$ in both parameterizations, since $q_\ell = P_\ell x_\ell$ is rank-$r$ regardless of where the kernel sits. 
For edge \glspl{cnn}, where $M_{\mathrm{saved}} \gg M_{\mathrm{optim}}$ (see Figure~\ref{fig:memory_story}), the $k\times$ parameter increase is a subdominant cost paid to recover adaptation capacity at the same activation floor.

\paragraph{Initialization.}
Because $P_\ell$ is frozen, its initialization fixes the subspace in which the adapter operates and the per-channel scale of the bottleneck. We initialize
$P_\ell \sim \mathcal{N}(0,1/r)$, for which
\begin{equation}
\mathbb{E}\!\left[\mathrm{Var}(q_{\ell,j})\right]
=
\tfrac{1}{r}\,\mathrm{tr}(\Sigma_{x_\ell}).
\end{equation}
This expected variance is identical across the $r$ bottleneck channels, where $\Sigma_{x_\ell}=\mathrm{Cov}(x_\ell)$.
This produces a variance-balanced bottleneck compatible with the scale-matched adapter in~\eqref{eq:final}. The trainable up-projection is zero-initialized, $U_{\ell}=0$, so the adapter is inactive at initialization and adaptation starts exactly from the source model. The scale mismatch of Section~\ref{sec:scale} therefore acts on the gradient rather than the forward pass.
We examine sensitivity to this initialization in Section~\ref{sec:results}.

\begin{table*}[t]
\centering
\caption{Headline Macro-F1 comparison at 50 adaptation steps. Mean $\pm$ sample standard deviations across 20 seeds are reported.}
\label{tab:headline_50_steps_std}
\footnotesize
\setlength{\tabcolsep}{2.0pt}
\renewcommand{\arraystretch}{0.82}
\resizebox{\textwidth}{!}{%
\begin{tabular}{llrcccccccc}
\toprule
Dataset & Backbone & Zero Shot & Full FT & Bias-Tuning & BN-Tuning & $r$ & LoRA-C & LoRA-Edge & MemFLoRA  & MemFLoRA-SG  \\
\midrule

\multirow{6}{*}{Opportunity}
& \multirow{3}{*}{MobileNetV2}
& & & & & 2
& 0.698 $\pm$ 0.037
& 0.730 $\pm$ 0.032
& 0.747 $\pm$ 0.025
& 0.758 $\pm$ 0.028 \\

& & 0.585 $\pm$ 0.088
& 0.820 $\pm$ 0.025
& 0.632 $\pm$ 0.054
& 0.668 $\pm$ 0.043
& 4
& 0.728 $\pm$ 0.032
& 0.760 $\pm$ 0.027
& 0.764 $\pm$ 0.028
& 0.774 $\pm$ 0.024 \\

& & & & & & 8
& 0.756 $\pm$ 0.028
& 0.789 $\pm$ 0.025
& 0.771 $\pm$ 0.028
& 0.785 $\pm$ 0.023 \\

\cmidrule(lr){2-11}

& \multirow{3}{*}{T-ResNet}
& & & & & 2
& 0.746 $\pm$ 0.044
& 0.710 $\pm$ 0.052
& 0.768 $\pm$ 0.046
& 0.773 $\pm$ 0.040\\

& & 0.511 $\pm$ 0.087
& 0.847 $\pm$ 0.026
& 0.655 $\pm$ 0.054
& 0.684 $\pm$ 0.052
& 4
& 0.773 $\pm$ 0.039
& 0.746 $\pm$ 0.046
& 0.793 $\pm$ 0.034
& 0.797 $\pm$ 0.035 \\

& & & & & & 8
& 0.801 $\pm$ 0.030
& 0.774 $\pm$ 0.038
& 0.812 $\pm$ 0.033
& 0.815 $\pm$ 0.031 \\

\midrule

\multirow{6}{*}{RealDisp}
& \multirow{3}{*}{MobileNetV2}
& & & & & 2
& 0.882 $\pm$ 0.005
& 0.892 $\pm$ 0.004
& 0.891 $\pm$ 0.004
& 0.896 $\pm$ 0.004 \\

& & 0.791 $\pm$ 0.000
& 0.894 $\pm$ 0.013
& 0.835 $\pm$ 0.003
& 0.865 $\pm$ 0.005
& 4
& 0.890 $\pm$ 0.005
& 0.898 $\pm$ 0.004
& 0.888 $\pm$ 0.010
& 0.897 $\pm$ 0.006 \\

& & & & & & 8
& 0.897 $\pm$ 0.003
& 0.903 $\pm$ 0.006
& 0.885 $\pm$ 0.010
& 0.895 $\pm$ 0.007 \\

\cmidrule(lr){2-11}

& \multirow{3}{*}{T-ResNet}
& & & & & 2
& 0.738 $\pm$ 0.009
& 0.631 $\pm$ 0.005
& 0.742 $\pm$ 0.017
& 0.768 $\pm$ 0.015 \\

& & 0.526 $\pm$ 0.000
& 0.891 $\pm$ 0.009
& 0.599 $\pm$ 0.005
& 0.620 $\pm$ 0.005
& 4
& 0.787 $\pm$ 0.010
& 0.686 $\pm$ 0.004
& 0.787 $\pm$ 0.010
& 0.815 $\pm$ 0.009 \\

& & & & & & 8
& 0.827 $\pm$ 0.010
& 0.742 $\pm$ 0.006
& 0.828 $\pm$ 0.011
& 0.847 $\pm$ 0.011 \\

\midrule

\multirow{6}{*}{RealWorld}
& \multirow{3}{*}{MobileNetV2}
& & & & & 2
& 0.802 $\pm$ 0.044
& 0.817 $\pm$ 0.046
& 0.874 $\pm$ 0.038
& 0.882 $\pm$ 0.035\\

& & 0.610 $\pm$ 0.087
& 0.911 $\pm$ 0.026
& 0.751 $\pm$ 0.067
& 0.825 $\pm$ 0.049
& 4
& 0.846 $\pm$ 0.037
& 0.866 $\pm$ 0.035
& 0.886 $\pm$ 0.035
& 0.892 $\pm$ 0.034 \\

& & & & & & 8
& 0.880 $\pm$ 0.032
& 0.889 $\pm$ 0.034
& 0.896 $\pm$ 0.034
& 0.901 $\pm$ 0.032 \\

\cmidrule(lr){2-11}

& \multirow{3}{*}{T-ResNet}
& & & & & 2
& 0.689 $\pm$ 0.096
& 0.632 $\pm$ 0.108
& 0.826 $\pm$ 0.065
& 0.847 $\pm$ 0.052 \\

& & 0.449 $\pm$ 0.124
& 0.903 $\pm$ 0.028
& 0.554 $\pm$ 0.134
& 0.631 $\pm$ 0.125
& 4
& 0.779 $\pm$ 0.067
& 0.717 $\pm$ 0.085
& 0.859 $\pm$ 0.047
&0.871 $\pm$ 0.044 \\

& & & & & & 8
& 0.847 $\pm$ 0.043
& 0.799 $\pm$ 0.056
& 0.875 $\pm$ 0.045
& 0.883 $\pm$ 0.040 \\

\bottomrule
\end{tabular}%
}

\end{table*}

\subsection{Projection-Gradient Recovery}
\label{sec:sg}

MemFLoRA freezes $P_\ell$ to preserve the memory floor, but this restricts adaptation to a fixed projected subspace.
As an optional extension, MemFLoRA-SG uses rematerialization, an execution-level technique orthogonal to the adapter design, to update $P_\ell$ without persistently retaining
the full-width input $x_\ell$.
Let $g_\ell=\partial\mathcal{L}/\partial q_\ell$, where $q_\ell=P_\ell x_\ell$. Then
\begin{equation}
\frac{\partial\mathcal{L}}{\partial P_\ell}
=
\sum_{b,t} g_{\ell,b,t}x_{\ell,b,t}^{\top}.
\end{equation}
Standard backpropagation retains $x_\ell$ until this outer product is formed.
MemFLoRA-SG stores only the rank-$r$ gradient $g_\ell$, replays the current adapted network under \texttt{no\_grad}, and immediately contracts
each rematerialized $x_\ell$ with $g_\ell$
~\cite{Chen2016TrainingDN}.
The input is released immediately after its layerwise contraction, so full-width
activations do not accumulate across layers and only one such buffer is live at a time. MemFLoRA-SG
therefore preserves rank-bounded persistent backward state,
$O(B\,r\,T_\ell)$, at the cost of one additional forward pass and a
layer-local full-width temporary buffer included in
the reported peak-memory accounting.

\section{Experimental Results}

\label{sec:results}

We follow the experimental protocol and application scope established by LoRA-Edge \cite{kwak2025loraedge}, the most directly related on-device CNN adaptation baseline, while extending the evaluation across three ranks and 20 independent seeds, with detailed training-state memory accounting and additional rank-independent baselines. We evaluate T-ResNet~\cite{bb_tresnet} and MobileNetV2~\cite{bb_mobilenetv2} on the three HAR benchmarks summarized in Table~\ref{tab:datasets}, spanning complementary
deployment shifts. Opportunity uses \gls{loso}, with one subject as target and the others as source; RealWorld uses \gls{lolo}, with one sensor location as target; and RealDisp uses ideal placement as source and self placement as target. Target data are divided into 80\% adaptation and 20\% held-out test data, with no test data used for adaptation or AdaBN calibration.

\begin{table}[t]
\centering
\caption{Datasets, domain shifts, and preprocessing settings.
}
\label{tab:datasets}

\setlength{\tabcolsep}{1.5pt}
\renewcommand{\arraystretch}{0.82}
\resizebox{0.83\columnwidth}{!}{%
\begin{tabular}{@{}llcccc@{}}
\toprule
Dataset & Shift & Subj. & Act. & Hz & Win./Str. \\
\midrule
Opportunity~\cite{ds_opportunity}
  & Subject & 4 & 17 & 30 & 60/30 \\
RealWorld~\cite{ds_realworld}
  & Body location & 15 & 8 & 50 & 500/250 \\
RealDisp~\cite{ds_realdisp}
  & Sensor placement & 17 & 33 & 50 & 250/125 \\
\bottomrule
\end{tabular}
}
\end{table}

All methods use Adam \cite{Kingma2014AdamAM} with a learning rate of $0.001$, a batch size of 64, and 50 adaptation steps to match the evaluation protocol of LoRA-Edge \cite{kwak2025loraedge}.
We adapt all convolutional layers in T-ResNet and the pointwise convolutions in MobileNetV2, and evaluate ranks $r\in\{2,4,8\}$. We report macro-F1 over 20 seeds. For each random seed, the source model is independently pretrained for 20 epochs; all adaptation runs use a weight decay of $0.0005$ and no learning-rate scheduler. MemFLoRA-SG replay uses the same mini-batch and unchanged parameters and completes before the optimizer step.

We compare against both rank-independent and rank-dependent adaptation baselines.
The rank-independent baselines are \textbf{Zero-shot}, which performs no target adaptation, \textbf{Full Fine-Tuning (Full FT)} which updates the entire model, \textbf{Bias-tuning}~\cite{ben-zaken-etal-2022-bitfit} and \textbf{BN-tuning}.
The rank-dependent baselines are \textbf{LoRA-C}~\cite{Ding2024LoRACPF} and \textbf{LoRA-Edge}~\cite{kwak2025loraedge}, two CNN-oriented LoRA
baselines for edge adaptation.

\subsection{Comparative Memory Analysis}



Table~\ref{tab:memory_rank_batch} reports training-memory
scaling at \(r=2\) across batch sizes for T-ResNet and MobileNetV2,
while Figure~\ref{fig:memory_story} visualizes the \(B=64\) peak-memory decomposition. Although the other methods reduce the number of trainable
parameters, their memory remains dominated by saved activations.
At \(B=64\), MemFLoRA reduces saved-activation memory relative to Full FT by
\(98.5\%\) on T-ResNet and \(98.7\%\) on MobileNetV2, with corresponding
peak training-state reductions of \(94.9\%\) and \(97.3\%\). 
MemFLoRA-SG retains rank-bounded persistent state but incurs a replay
buffer, yielding peak-memory reductions of \(92.1\%\) and \(96.1\%\),
respectively. For fairness, all rank-dependent methods use
activation-aware implementations that retain only tensors required by their adapter gradients. The batch-size sweep further shows that
MemFLoRA's saved-state advantage persists as activation memory grows. Notably, its peak memory at \(B=64\), \(3.23\) MB on T-ResNet and
\(17.83\) MB on MobileNetV2, is comparable to LoRA-Edge at \(B=1\),
with \(2.93\) MB and \(17.09\) MB, respectively. Thus, under a similar
memory budget, MemFLoRA supports a \(64\times\) larger adaptation batch.

\begin{table}[t]
\centering
\caption{For T-ResNet and MobileNetV2, batch-size entries report
peak/saved-activation memory [MB]; optimizer-state memory is reported
separately.}
\label{tab:memory_rank_batch}

\setlength{\tabcolsep}{1.25pt}
\renewcommand{\arraystretch}{1}
\resizebox{\columnwidth}{!}{%
\begin{tabular}{@{}clccccc@{}}
\toprule
& Method
& $B_1$ & $B_8$ & $B_{32}$ & $B_{64}$
& Opt. [MB] \\
\midrule

\multirow{5}{*}{\rotatebox[origin=c]{90}{T-ResNet}}
& Full FT
& 9.02/0.90
& 13.84/7.09
& 35.08/28.33
& 63.40/56.65
& 4.49 \\

\cmidrule(l){2-7}

& LoRA-C
& 5.51/3.10
& 11.53/9.12
& 32.20/29.79
& 59.76/57.35
& 0.10 \\

& LoRA-Edge
& 2.93/0.60
& 7.05/4.72
& 21.19/18.86
& 40.04/37.71
& 0.02 \\

& MemFLoRA
& \textbf{2.44/0.02}
& \textbf{2.50/0.11}
& \textbf{2.81/0.42}
& \textbf{3.23/0.83}
& 0.08 \\

& MemFLoRA-SG
& 2.50/0.02
& 2.78/0.11
& 3.74/0.42
& 5.01/0.83
& 0.10 \\

\midrule

\multirow{5}{*}{\rotatebox[origin=c]{90}{MobileNetV2}}
& Full FT
& 43.20/16.12 & 109.00/81.91 & 346.93/319.85 & 666.64/639.56 & 17.97 \\

\cmidrule(l){2-7}

& LoRA-C & 27.17/17.61 & 90.00/80.44 & 305.40/295.84 & 592.61/583.05 & 0.30 \\

& LoRA-Edge & 17.09/7.66 & 69.77/60.34 & 250.39/240.96 & 491.22/481.79 & 0.16 \\

& MemFLoRA & \textbf{9.63/0.20} & \textbf{10.54/1.11} & \textbf{13.66/4.23} & \textbf{17.83/8.40} & 0.16 \\

& MemFLoRA-SG & 9.91/0.20 & 11.73/1.11 & 17.97/4.23 & 26.30/8.40 & 0.16 \\

\bottomrule
\end{tabular}%
}
\end{table}

\subsection{Results Across Adaptation Budgets}
\label{sec:results-accuracy}

Table~\ref{tab:headline_50_steps_std} reports macro-F1 under a tight 50-step adaptation budget across six dataset--backbone settings and three ranks. Among the single-pass rank-dependent methods, MemFLoRA reaches an average of $0.827$, compared with $0.798$ for LoRA-C and $0.777$ for LoRA-Edge. Thus, its substantial memory reduction does not come at the cost of adaptation quality. MemFLoRA-SG provides an accuracy-oriented operating point by increasing the average macro-F1 to $0.839$, while retaining a substantially smaller memory footprint than conventional CNN LoRA methods.

\begin{figure}[tb!]
    \centering
    \includegraphics[width=\linewidth]{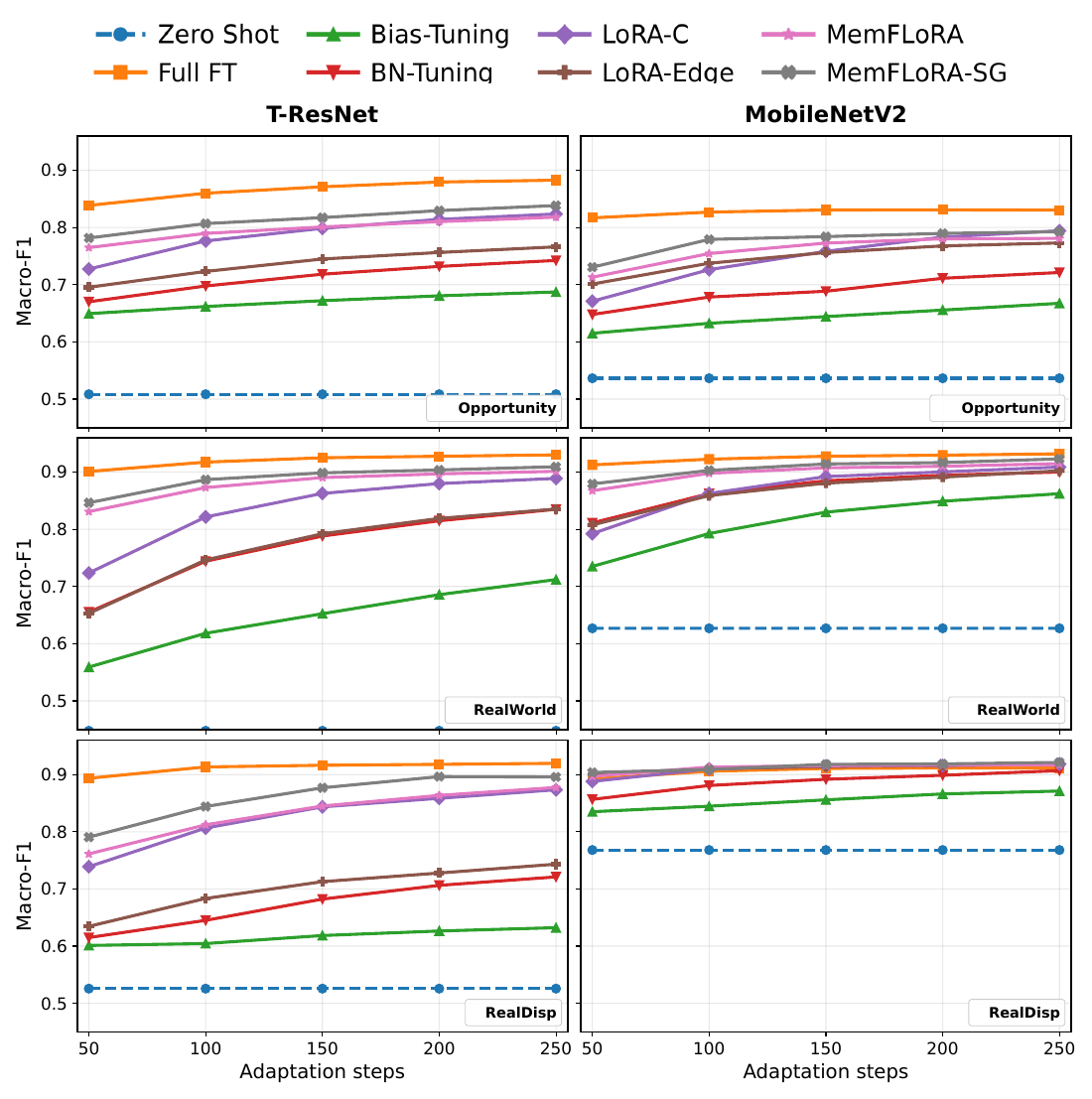}
    \caption{Adaptation performance at $rank=2$ across adaptation steps for T-ResNet (left) and MobileNetV2 (right).}
    \label{fig:tresnet_rank2_rank4_sharedlegend}
\Description[Adaptation trajectories across datasets and backbones]{
Six line charts show the macro-averaged F1 score as the number of
adaptation steps increases from fifty to two hundred fifty. The left
column presents T-ResNet and the right column presents MobileNetV2.
The rows present Opportunity, RealWorld, and RealDisp, respectively.
Each panel compares Zero Shot, Full Fine-Tuning, Bias-Tuning,
Batch-Normalization Tuning, LoRA-C, LoRA-Edge, MemFLoRA, and
MemFLoRA-SG. Zero Shot appears as a constant dashed baseline, while
the adapted methods generally improve during the first adaptation
checkpoints and then level off. Full Fine-Tuning is highest or close
to highest across the panels. MemFLoRA-SG is generally the strongest
low-rank method and is closely followed by MemFLoRA. The RealWorld
T-ResNet panel shows the widest separation among methods, whereas
the MobileNetV2 trajectories are more tightly clustered,
particularly on RealDisp.
}
\end{figure}

We next examine convergence across adaptation budgets. Figure~\ref{fig:tresnet_rank2_rank4_sharedlegend} shows the rank-2 adaptation trajectories, which are qualitatively consistent across datasets and backbones.
MemFLoRA narrows the gap to Full FT as adaptation proceeds while remaining competitive at convergence. MemFLoRA-SG generally provides the strongest performance through projection-gradient recovery.

\subsection{Design Ablations and Training Cost}
Table~\ref{tab:ablation_design_choices} summarizes three design choices. 
First, Gaussian initialization consistently outperforms the orthogonalized projection across the three evaluated ranks.
We therefore use Gaussian initialization in the remaining experiments.
Second, placing the spatial kernel in the trainable factor improves macro-F1 while preserving the same rank-bounded saved-state.
Finally, the scale-normalization ablation shows that adding an unnormalized adapter to the normalized backbone output causes performance to collapse.
Either bottleneck normalization or the frozen source-BN scale restores a usable adapter scale, but the source-BN scale alone gives the best average result. 
Following prior on-device adaptation protocols, we calibrate AdaBN
using one batch of target data, showing that it has a negligible effect. Entries in the table report mean macro-F1 over all target subjects from an independent ablation sweep; differences from Table \ref{tab:headline_50_steps_std}  are within seed-level variation.

\begin{table}[tb]
\centering
\setlength{\tabcolsep}{3pt}
\renewcommand{\arraystretch}{0.9}
\caption{Ablations on Opportunity with T-ResNet at 50 adaptation steps.}
\label{tab:ablation_design_choices}
\resizebox{0.65\columnwidth}{!}{%
\begin{tabular}{llccc}
\toprule
Factor & Variant & $r=2$ & $r=4$ & $r=8$ \\
\midrule
Init.
& Normal $P$
& 0.759 & 0.777 & 0.802 \\
& Orthogonal $P$
& 0.673 & 0.709 & 0.751 \\
\midrule
Geom.
& $1{\times}1$ down, $k{\times}k$ up
& 0.760 & 0.777 & 0.802 \\
& $k{\times}k$ down, $1{\times}1$ up
& 0.731 & 0.755 & 0.773 \\
\midrule
Scale
& None
& 0.200 & 0.197 & 0.187 \\
& $BN_r$
& 0.728 & 0.772 & 0.805 \\
& $a_s$
& \textbf{0.764} & \textbf{0.792} & \textbf{0.808} \\
\midrule
AdaBN
& Off
& 0.753 & 0.779 & 0.804 \\

& On
& 0.762 & 0.783 & 0.801 \\

\bottomrule
\end{tabular}
}

\end{table}



Table~\ref{tab:training_perf_rank2_compact_adapt} summarizes the training-side cost at rank 2 on Opportunity. MemFLoRA closely matches the computational cost of LoRA-Edge:
$222.5$ versus $218.6$ million MACs for T-ResNet and $286.0$ versus
$283.2$ million MACs for MobileNetV2. These correspond to only 70\%
and 71\% of the Full Fine-Tuning cost, respectively. MemFLoRA uses
more trainable parameters than LoRA-Edge on T-ResNet
($10.6$K versus $2.3$K) and the same number on MobileNetV2
($19.8$K). This is a deliberate geometry-level trade-off: the
trainable parameter footprint remains small, while the adapter
directly reduces the activations saved for backward, which dominate CNN
training memory. MemFLoRA therefore provides comparable or better
adaptation quality and nearly identical computational cost to LoRA-Edge, while
substantially reducing peak training-state memory. MemFLoRA-SG retains the same number of trainable parameters
but requires 104\% and 106\% of the Full Fine-Tuning MACs because of
its additional forward replay. It therefore offers higher adaptation
accuracy in exchange for additional computation, while remaining in
the same low-memory regime as MemFLoRA. 
To isolate the convergence behavior during adaptation, we measure the
adaptation steps until each method first reaches 85\% of the final Full FT
macro-F1 at the adaptation checkpoints shown in Figure~\ref{fig:tresnet_rank2_rank4_sharedlegend}. MemFLoRA and MemFLoRA-SG reach the threshold fastest on T-ResNet and are comparable to LoRA-Edge on MobileNetV2.

\begin{table}[tb]
\centering
\setlength{\tabcolsep}{3pt}
\renewcommand{\arraystretch}{0.9}
\caption{Training-performance summary at rank 2. Total adaptation
\glspl{mac} are reported as AdaBN calibration \glspl{mac} plus 50
adaptation steps.}
\label{tab:training_perf_rank2_compact_adapt}
\resizebox{0.82\columnwidth}{!}{%
\begin{tabular}{llrrrr}
\toprule
Backbone & Method
& \shortstack{Par.\\$[{10^3}]$}
& \shortstack{\glspl{mac}\\$[{10^6}]$}
& \shortstack{Relative\\cost $[\%]$}
& \shortstack{Steps to\\$85\%$ $[n]$} \\
\midrule

\multirow{5}{*}{T-ResNet}
& Full FT
& 560.9 & 320.0 & -- & -- \\

& LoRA-C
& 12.8 & 320.2 & 100 & 22 \\

& LoRA-Edge
& \textbf{2.3} & \textbf{218.6} & \textbf{68} & 24\\

& MemFLoRA
& 10.6 & 222.5 & 70 & 19 \\

& MemFLoRA-SG
& 10.6 & 331.7 & 104 & \textbf{15}\\

\midrule

\multirow{5}{*}{MobileNetV2}
& Full FT
& 2250 & 403.0 & -- & -- \\

& LoRA-C
& 37.1 & 390.9 & 97 & 32 \\

& LoRA-Edge
& \textbf{19.8} & \textbf{283.2} & \textbf{70} & 24 \\

& MemFLoRA
& \textbf{19.8} & 286.0 & 71 & 24\\

& MemFLoRA-SG
& \textbf{19.8} & 426.0 & 106 & \textbf{23} \\

\bottomrule
\end{tabular}%
}
\end{table}


MemFLoRA also outperforms non-\gls{lora} activation-minimal approaches, such as TinyTL~\cite{NEURIPS2020_81f7acab}.
Since TinyTL is not a plug-and-play \gls{lora} adapter, its implementation requires architecture-specific design choices; we therefore evaluate both an activation-minimal frozen-block variant with eval-mode BN and a train-BN variant that preserves BN adaptation at higher memory cost. On Opportunity, these variants reach 0.682 and 0.737 macro-F1 with 150.47 MB and 518.83 MB peak training-state memory, respectively. In contrast, rank-2 MemFLoRA reaches 0.747 macro-F1 with only 17.83 MB, showing that our low-rank memory-floor design provides a simpler adapter-style alternative with substantially lower training memory.




\subsection{Jetson Orin Deployment Validation}

We validate training-state memory on an NVIDIA Jetson Orin Nano
(8\,GB) using both backbones and all methods at $r=2$ and $B=64$.
We measure one complete update after two warm-up updates using Opportunity-shaped inputs.
The device profiler tracks allocations associated
with model parameters and buffers, materialized optimizer state,
gradients, saved-backward state, and total training state.
Peak memory denotes the maximum concurrent storage of these tracked
training-state components, whereas saved-backward state is measured
separately at the end of the forward pass. Pre-existing inputs and
labels, checkpoint copies, general intermediates and workspaces, and allocator cache are outside this metric.

\begin{table}[tb]
\centering
\setlength{\tabcolsep}{3pt}
\caption{Training-state memory measured at
rank $r=2$ and batch size $B=64$. Entries report peak concurrent
training-state storage / saved-backward state [MB].}
\label{tab:jetson_memory}

\begin{tabular}{lcc}
\toprule
Method & T-ResNet & MobileNetV2 \\
\midrule
Full FT       & 54.52 / 47.77 & 666.72 / 639.55 \\
LoRA-C        & 52.37 / 49.96 & 592.61 / 583.04 \\
LoRA-Edge     & 35.12 / 32.79 & 491.21 / 481.78 \\
MemFLoRA      &  3.22 /  0.83 &  17.83 /   8.39 \\
\bottomrule
\end{tabular}

\end{table}

At $B=64$, MemFLoRA reduces saved-backward storage relative to
Full FT by 98.3\% and 98.7\% on T-ResNet and MobileNetV2,
respectively, with corresponding peak training-state reductions of
94.1\% and 97.3\%. These measurements closely follow the memory
reduction trend estimated in Table~\ref{tab:memory_rank_batch}, where the
corresponding reductions were 98.5\%/98.7\% for saved state and
94.9\%/97.3\% for peak training state. 
Nevertheless, the deployment measurements preserve the large training-state memory advantage of MemFLoRA.

Table~\ref{tab:jetson_runtime} reports the peak requested CUDA
memory from the same runs. This footprint includes training
state, input batches, transient tensors, and workspaces, but
excludes unused allocator cache and allocation rounding.
These additional runtime allocations lie outside the
training-state memory floor of Section~\ref{sec:floor}.

\begin{table}[tb]
\centering
\setlength{\tabcolsep}{3pt}
\caption{Peak requested CUDA memory at
$r=2$, $B=64$ [MB], from the same runs as
Table~\ref{tab:jetson_memory}. Values include inputs,
temporaries, and workspaces in addition to training state.}
\label{tab:jetson_runtime}
\setlength{\tabcolsep}{6pt}

\begin{tabular}{@{}lrr@{}}
\toprule
Method & T-ResNet & MobileNetV2 \\
\midrule
Full FT     & 66.20 & 673.39 \\
LoRA-C      & 64.08 & 601.45 \\
LoRA-Edge   & 44.80 & 503.32 \\
MemFLoRA    & 19.43 &  95.43 \\
\bottomrule
\end{tabular}
\end{table}

\begin{figure}[tb]
\centering
\includegraphics[width=\columnwidth]{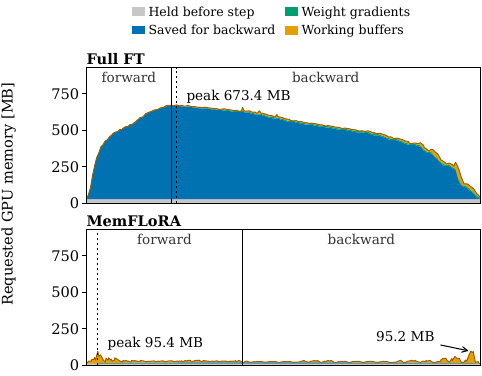}
\caption{Requested GPU memory during
one MobileNetV2 update at $r=2$, $B=64$ on Jetson Orin Nano.}
\label{fig:jetson_timeline}
\Description{Two stacked area plots of requested GPU memory over one
training update. Full FT rises during the forward pass to a peak of
673.4 MB dominated by saved activations, then declines during the
backward pass. MemFLoRA stays low, with a narrow 95.4 MB spike of
working buffers early in the forward pass and a second 95.2 MB spike
at the end of the backward pass.}
\end{figure}

Figure~\ref{fig:jetson_timeline} traces the requested memory through the
forward and backward passes of one MobileNetV2 update at $B=64$,
recorded in a separate Full FT and MemFLoRA run; the optimizer step
adds no peak and is omitted. Full FT peaks at the start of the backward pass, when
639.2\,MB of saved activations are still held. MemFLoRA instead peaks
early in the forward pass. The first, high-resolution layers produce
the largest feature maps,
and their transient intermediates occupy 82.7 of MemFLoRA's
95.4\,MB peak, while only 1.5\,MB is saved for backward. These buffers
are released after each layer instead of accumulating. A second,
nearly equal spike (95.2\,MB) occurs at the end of the backward pass,
when backpropagation reaches the same early layers and their
full-width activation gradients are formed. Full FT shows the same late
spike in working buffers, but by then most of its saved
activations have been released. MemFLoRA's peak is therefore set by
the transient working set of the largest layers, not by retained
state.
 
\section{Conclusion}
MemFLoRA shows that efficient edge CNN adaptation depends not only on trainable parameters but also on activations retained for backpropagation. It reduces saved-activation memory by 98.5--98.7\% and peak training-state memory by 94.9--97.3\% relative to full fine-tuning. By defining an activation-memory-floor criterion, deriving the design constraints it imposes, and showing that adaptation quality can be preserved while satisfying it, this work reframes LoRA for on-device CNN adaptation as a saved-state design problem rather than merely a parameter-count problem, providing a foundation for future edge adaptation research. 

\begin{acks}
This work has been developed in the project DI-EDAI funded
by the German Federal Ministry of Research, Technology and Space  (BMFTR) under contract No. 16ME0991.
\end{acks}

\clearpage
\balance
\bibliographystyle{ACM-Reference-Format}
\bibliography{sample-base}


\end{document}